\newcommand{\CLASSINPUTbottomtextmargin}{1in}
\documentclass[conference,letterpaper]{IEEEtran}
\ifCLASSINFOpdf
\else
\fi
\usepackage[T1]{fontenc}
\usepackage{graphicx}
\usepackage{amsmath}
\usepackage{amsfonts}
\usepackage{multirow}
\usepackage{subcaption}
\usepackage{booktabs}
\usepackage{algorithm}
\usepackage{algpseudocode}
\usepackage{setspace}
\begin{document}
%
% paper title
% can use linebreaks \\ within to get better formatting as desired
\title{An Empirical Study on Predictive Maintenance for Component X in Heavy-Duty Scania Trucks}

% author names and affiliations
% use a multiple column layout for up to three different
% affiliations
\author{\IEEEauthorblockN{Valeriu Dimidov, Sasan Jafarnejad and Rapha\"el Frank}
\IEEEauthorblockA{
Interdisciplinary Centre for Security, Reliability and Trust (SnT)\\
University of Luxembourg\\
29 Avenue J.F. Kennedy L-1855, Luxembourg\\
firstname.lastname@uni.lu}
}

% conference papers do not typically use \thanks and this command
% is locked out in conference mode. If really needed, such as for
% the acknowledgment of grants, issue a \IEEEoverridecommandlockouts
% after \documentclass

% for over three affiliations, or if they all won't fit within the width
% of the page, use this alternative format:
% 
%\author{\IEEEauthorblockN{Michael Shell\IEEEauthorrefmark{1},
%Homer Simpson\IEEEauthorrefmark{2},
%James Kirk\IEEEauthorrefmark{3}, 
%Montgomery Scott\IEEEauthorrefmark{3} and
%Eldon Tyrell\IEEEauthorrefmark{4}}
%\IEEEauthorblockA{\IEEEauthorrefmark{1}School of Electrical and Computer Engineering\\
%Georgia Institute of Technology,
%Atlanta, Georgia 30332--0250\\ Email: see http://www.michaelshell.org/contact.html}
%\IEEEauthorblockA{\IEEEauthorrefmark{2}Twentieth Century Fox, Springfield, USA\\
%Email: homer@thesimpsons.com}
%\IEEEauthorblockA{\IEEEauthorrefmark{3}Starfleet Academy, San Francisco, California 96678-2391\\
%Telephone: (800) 555--1212, Fax: (888) 555--1212}
%\IEEEauthorblockA{\IEEEauthorrefmark{4}Tyrell Inc., 123 Replicant Street, Los Angeles, California 90210--4321}}

% use for special paper notices
%\IEEEspecialpapernotice{(Invited Paper)}

% make the title area
\maketitle

\begin{abstract}
%\boldmath
Condition-based Predictive Maintenance (PdM) for truck fleets has gained momentum in recent years. 
This maintenance strategy aims to minimize unplanned downtimes and reduce costs by monitoring the health status of vehicles and taking proactive action based on their condition.
However, the implementation of condition-based PdM systems is challenging due to the large volume of data generated by the trucks, the inherent complexity of detecting failures through sensor data and the difficulties in finding cost-effective trade-offs in the solution’s implementation. 
In this paper, we define and validate a condition-based PdM methodology built on the assumption that the wear-and-tear state of the monitored component can be represented as a monotonically non-decreasing time series. 
It involves selecting only the most recent observations from the time series and transforming them into a tabular format for classification using machine learning (ML) models designed for tabular data. 
Our results indicate that the proposed methodology reduces costs on the Scania Component X dataset compared to current state-of-the-art (SOTA) approaches, while also simplifying the modeling process through AutoML.
\end{abstract}


% IEEEtran.cls defaults to using nonbold math in the Abstract.
% This preserves the distinction between vectors and scalars. However,
% if the conference you are submitting to favors bold math in the abstract,
% then you can use LaTeX's standard command \boldmath at the very start
% of the abstract to achieve this. Many IEEE journals/conferences frown on
% math in the abstract anyway.

% no keywords

% For peer review papers, you can put extra information on the cover
% page as needed:
% \ifCLASSOPTIONpeerreview
% \begin{center} \bfseries EDICS Category: 3-BBND \end{center}
% \fi
%
% For peerreview papers, this IEEEtran command inserts a page break and
% creates the second title. It will be ignored for other modes.
\IEEEpeerreviewmaketitle

\section{Introduction}
\label{sec:introduction}
% no \IEEEPARstart
Conventional automobile maintenance strategies consist of either fixing a vehicle post-failure or conducting pre-scheduled technical checks to service a system before a malfunction occurs \cite{theissler_predictive_2021}. 
Although these methods have supported the industry for decades \cite{pintelon_2008}, they have some drawbacks in terms of costs and efficiency. 
The first strategy, referenced as reactive maintenance (RM), is prone to unexpected downtime and consequently expensive repairs as it passively waits for failures to happen. 
The second one is known as preventive maintenance (PM). It is a proactive strategy since it tries to anticipate the failure using static time and mileage rules. 
However, it lacks flexibility and poses challenges in estimating the ideal inspection frequency. Adopting a conservative approach limits the maximization of component utilization and results in unnecessary checks. 
In contrast, an approach with fewer pre-scheduled maintenance sessions risks missing early signs of failures, sometimes causing PM to degenerate into RM.

In recent years, technological advances have paved the way for predictive maintenance (PdM) in the automotive sector \cite{killeen_2019}.
Condition-based PdM \cite{theissler_predictive_2021} involves monitoring the health state of vehicle components through integrated sensors to detect when they approach failure.
This is usually achieved using a telematics gateway connected to the vehicle’s Controller Area Network (CAN) bus. 
The device collects data from a range of sensors, including parameters such as fuel consumption, engine speed, and Diagnostic Trouble Codes (DTC).
The collected data is then transmitted to a centralized repository for storage, preprocessing and analysis. 
Since the relationship between system health and failure is complex, ML models are employed to estimate the remaining useful life (RUL) of the components.
Alerts to fleet managers are triggered when models detect symptoms of imminent failures, signaling the need for maintenance.
 
Nevertheless, large-scale adoption of PdM in the automotive industry faces several challenges. 
The authors of \cite{theissler_predictive_2021} highlight many technical obstacles, including the limited availability of real datasets which restricts open and reproducible research,  the intrinsic complexity of PdM problems like heterogeneity of fleets, rarity of faults and resilience to adopt ML-driven maintenance due to concerns over model interpretability and data privacy. 
Similarly, in another literature review \cite{arena_predictive_2022}, the authors identify the initial capital for equipment acquisition and the need for employee training as the main barriers to start business activities related to PdM.

To address these challenges, particularly the concerns over model interpretability and implementation costs highlighted in \cite{theissler_predictive_2021} and \cite{arena_predictive_2022}, this work proposes a simplified yet effective approach to PdM.
Our methodology reduces both the technical complexity and computational overhead typically associated with time-series analysis, while maintaining competitive performance compared to more complex solutions.

\textbf{Contribution}: This research introduces a new methodology for approximating time-series models for condition-based PdM applications.  
Under the assumption that the wear-and-tear state of the monitored component is represented as a monotonically non-decreasing time series, the proposed methodology focuses on leveraging the last observation of the time series to train tabular ML models.
The methodology demonstrates improved performance over current SOTA \cite{carpentier, parton, zhong} results reported in the IDA Challenge 2024, at the same time it simplifies model training through the use of an AutoML tools.

The structure of the paper is organized as follows. 
Section 2 presents a review of relevant literature. 
In Section 3, we define the problem, provide an overview of the dataset, describe the evaluation metrics, and present the AutoML and SHAP frameworks utilized in this study.
In Section 4 we present the proposed methodology and detail the experimental setup. 
Section 5 reports and analyzes the experimental results, comparing these findings with SOTA approaches.
Section 6 focuses on the explainability of our model. 
Section 7 discusses the main strengths and limitations of our methodology. 
Finally, in the last section, we suggest potential directions for future research.
% You must have at least 2 lines in the paragraph with the drop letter
% (should never be an issue)
\section{Related Works}
\label{sec:related_works}

Recent industrial challenges held at scientific conferences have accelerated the dissemination of PdM methodologies in automotive sector, particularly in the domain of time-series modeling. 
The IDA 2024 Industrial Challenge tasked participants with developing predictive models to identify imminent failures of Component X in heavy-duty Scania trucks. 

In this context, the methodology proposed by \cite{carpentier} involved extracting features from operational readouts using various window-based strategies combined with the tsfresh library. 
Feature selection was performed using Kendall’s Tau, after which a range of models—including classification, regression, and survival analysis—were applied to predict the degradation class and RUL. 

The work presented in \cite{parton} introduces an innovative approach by converting time series data into graphs using the visibility graph algorithm and subsequently applying Graph Neural Networks (GNN) to classify the time series into two classes (healthy and faulty). 
The methodology involves identifying suitable univariate time series for graph representation by performing Principal Component Analysis (PCA) on a similarity matrix derived from the truncated signature of each series. 
Subsequently, Graph Isomorphism Network (GIN) \cite{xu_how_2019} was used to perform binary classification on the multivariate time series and vehicle specification.

The authors of \cite{zhong} focused on deep learning techniques, implementing different types of Multilayer Perceptrons (MLP), Convolutional Neural Networks (CNN), and Recurrent Neural Networks (RNN) to predict failures.
So far their methodology has achieved the best results in terms of the evaluation metric proposed in the challenge.

Similar approaches have been devised for estimating RUL of turbofan jet engines utilizing the well-known C-MAPSS dataset \cite{saxena_turbofan_2008}. For instance, \cite{shi_dual-lstm_2021} introduces a Dual-LSTM model designed to estimate system RUL by integrating change point identification with RUL forecasting via a health index (HI). In \cite{li_remaining_2018}, CNNs are utilized by the authors to automatically extract features from raw sensor data, eliminating the need for prior knowledge in prognostics or signal processing. Following this, a fully connected layer (FNN) is employed to estimate the RUL. Finally, \cite{listou_ellefsen_remaining_2019} propose a hybrid architecture combining unsupervised pre-training based on Restricted Boltzmann machine (RBM) with supervised learning (LSTM + FNN) to enhance feature extraction and improve RUL estimation.

\section{IDA 2024 Challenge}
\label{sec:experimental_setup}

This section formally defines the problem under investigation and provides a comprehensive overview of the dataset employed in this study. It also describes the evaluation metrics used to measure model performance and introduces the ML framework adopted for the analysis. Additionally, we elaborate on the integration of ensembling techniques into the model training process.

% \noindent TO-DO \\
\subsection{Problem Definition} \label{sec:problem_definition}

We consider a set of $n$ vehicles, denoted as $V = \{V_{0}, V_{1}, \dots, V_{n-1}\}$, monitored using onboard sensors. 
The monitored signals are post-processed to generate a multivariate time series $X_{i} \in \mathbb{R}^{T_{i} \times M}$, where $T_i \in \mathbb{N}$ denotes the number of observations for vehicle $V_i$, and $M$ represents the number of features. The time series \( X_{i} \) is assumed to be monotonically non-decreasing and irregularly sampled, capturing the progressive wear-and-tear state of the studied component.

Each observation \( x_{t} \in X_{i} \) is associated with a label \( y_{t} \in \{0, 1, 2, 3, 4\} \), which represents the degradation level of the monitored component at time step \( t \). 
The sequence of labels for vehicle \( V_i \) is denoted as \( Y_{i} \in \{0, 1, 2, 3, 4\}^{T_i} \). This sequence is also assumed to be monotonically non-decreasing and irregularly sampled, consistent with the characteristics of \( X_{i} \).

Given the dataset \( \mathcal{D} = \{ (X_i, Y_i) \}_{i=0}^{n-1} \), the objective is to learn a mapping function:
\begin{equation}
\label{eq:mapping_function}
    f_{TS}: \bigcup_{T \in \mathbb{N}} \mathbb{R}^{T \times M} \rightarrow \{0, 1, 2, 3, 4\}
\end{equation}

\noindent such that, for a given time series \( X_i \), the function \( f_{TS} \) predicts the degradation level of the last observation in the sequence.

\subsection{Dataset} \label{sec:dataset}

In our experiments, we used the Scania Component X dataset provided by \cite{kharazian_2024} for the industrial challenge. 
The dataset contains real-world multivariate time series collected from a fleet of 33,641 trucks. 
It is divided into training (70\%), validation (15\%), and test (15\%)  with respectively 23550, 5046 and 5045 vehicles. It includes vehicle specifications and repair records (off-board data) as well as operational readouts related to an anonymized engine component (on-board data). 

Next, we highlight the key aspects of the dataset. For more details we refer the reader to \cite{kharazian_2024}.
\subsubsection{Vehicle Specifications}
\label{sub:vehicle-specification}
The vehicle specification dataset consists of eight categorical features (named incrementally from Spec\_0 to Spec\_7), describing static attributes of each truck in the dataset, such as engine type, wheel configuration, and other key specifications.
\subsubsection{Time-to-Event}
\label{sec:tte}
The time-to-event dataset records maintenance activities related to Component X. 
For each vehicle, it specifies whether Component X experienced a failure during the study period (in\_study\_repair) and the time step at which the maintenance activity occurred (length\_of\_study\_time\_step). 
In particular, this information is available exclusively for vehicles in the training dataset.
In contrast, for the validation and test partitions, only the health class label associated with the last recorded operational readout is provided.
\subsubsection{Operational Readouts}
The operational readouts dataset is a panel of multivariate time-series, each corresponding to a specific vehicle. 
The time-series comprise 107 features categorized into two types: single-counter and histogram-based features. 
Single-counter features (8 in total) represent cumulative trends, while histogram-based features (6 sets) summarize operational conditions across multiple bins. Both types exhibit increasing trends, representing the health state of the component as a monotonic non-decreasing time series. 

Nevertheless, the post-processing routine applied to the time series does not account for all possible types of corruption, as noted in \cite{kharazian_2024}. In particular, we have identified downwards glitches and cumulative step noise.

\begin{figure}[H]
    \centering
    \begin{subfigure}[b]{\linewidth}
        \centering
        \includegraphics[width=\linewidth]{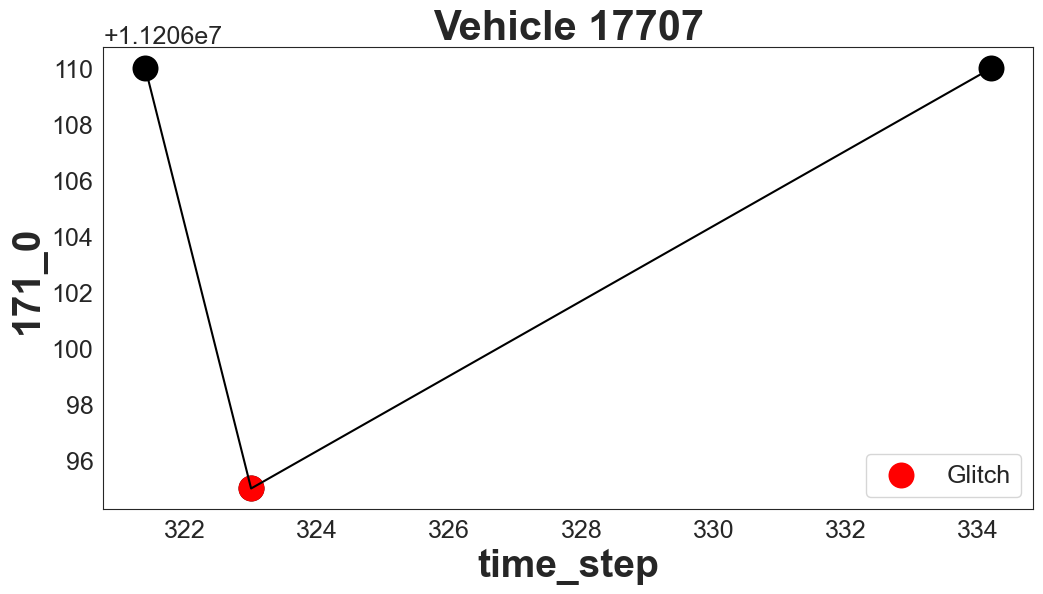}
        \caption{Glitch}
        \label{fig:glitch}
    \end{subfigure}
    \vfill
    \begin{subfigure}[b]{\linewidth}
        \centering
        \includegraphics[width=\linewidth]{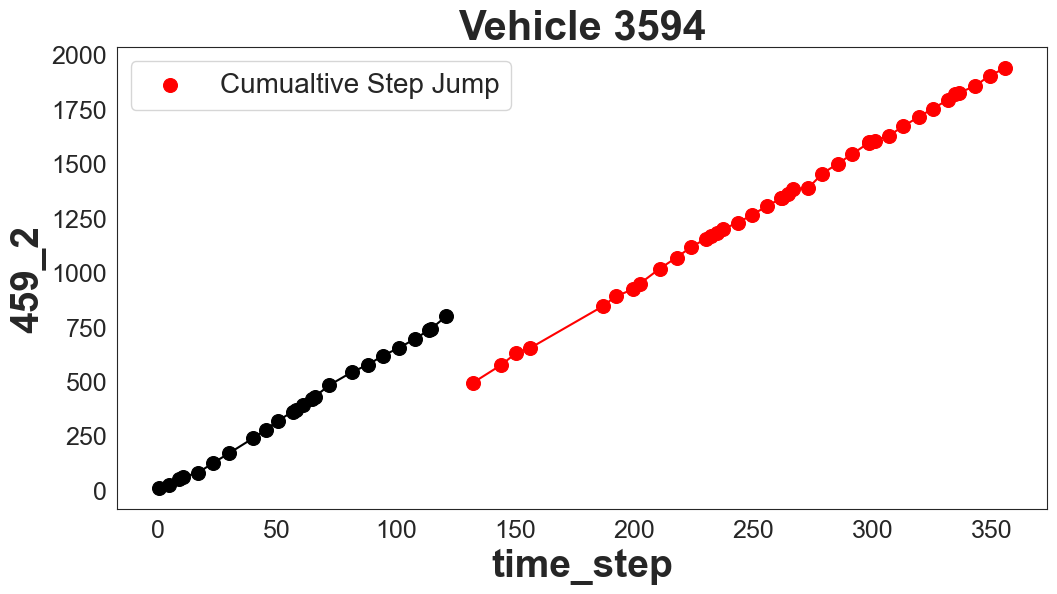}
        \caption{Cumulative step noise}
        \label{fig:downwards_step_jump}
    \end{subfigure}
    \caption{Types of noise in operational readouts}
    \label{fig:noise_types}
\end{figure}

Glitches (Figure \ref{fig:glitch}) manifest as sudden, transient deviations from normal behavior before quickly returning to expected levels. In contrast, cumulative downward step noise (Figure \ref{fig:downwards_step_jump}) is characterized by abrupt, persistent drops in measured values.

While glitches typically result from sensor errors or temporary system anomalies, cumulative step noise arises from various factors, such as an ECU software update altering recorded parameters or a temporary disconnection of the telematics device, leading to data loss or misalignment.

We also hypothesize the presence of upward noise. 
However, its identification remains challenging due to the monotonically non-decreasing nature of the time series. Additionally, alterations in time series trends can result from variations in vehicle usage.
\subsection{Performance Metrics} \label{sec:evaluation-metrics}

ML models for IDA 2024 challenge are typically evaluated using a custom metric introduced in  \cite{kharazian_2024}. 
This cost function assigns different penalties associated with different types of misclassifications.
In particular, it imposes higher penalties on false negatives, with the cost increasing proportionally to the severity of the component's degradation state.

\begin{table}[h]
     \caption{Cost Matrix. Source: \cite{kharazian_2024}}
    \centering
    \begin{tabular}{|c|c|c|c|c|c|c|}
        \hline
        & \multicolumn{6}{|c|}{\textbf{Predicted}} \\ \hline
        \multirow{6}{*}{\rotatebox{90}{\textbf{Actual}}} && \textbf{0} & \textbf{1} & \textbf{2} & \textbf{3} & \textbf{4} \\ \cline{2-7}
        & \textbf{0} & 0   & 7   & 8   & 9   & 10 \\ \cline{2-7}
        & \textbf{1} & 200 & 0   & 7   & 8   & 9  \\ \cline{2-7}
        & \textbf{2} & 300 & 200 & 0   & 7   & 8  \\ \cline{2-7}
        & \textbf{3} & 400 & 300 & 200 & 0   & 7  \\ \cline{2-7}
        & \textbf{4} & 500 & 400 & 300 & 200 & 0  \\ \hline
    \end{tabular}
    \label{tab:cost_matrix}
\end{table}

\noindent Let \textit{m} denote a ML model, let \textit{C} indicate the cost matrix outlined in Table \ref{tab:cost_matrix}, and let $\mathit{M_{ \mathcal{D}}}$ represent the confusion matrix generated by evaluating \textit{m} on the Scania dataset $ \mathcal{D}$.
The cost of \textit{m} for the Component X PdM problem is expressed as:

\begin{equation} \text{Cost($\mathit{M_{ \mathcal{D}}}$)} = \sum_{i=0}^{4} \sum_{j=0}^{4} C[i,j] \cdot M_{ \mathcal{D}}[i,j] 
\label{eq:challege-metric}
\end{equation}

\noindent where: 
\begin{itemize} 
    \item \( C[i,j] \) is the cost of predicting class \( j \) when the true class is \( i \)
    \item \( M_{ \mathcal{D}}[i,j] \) is the number of instances where the true class is \( i \) and the predicted class is \( j \)
\end{itemize}

\vspace{0.2cm}

\noindent In addition, we also consider the Balanced Accuracy metrics, since it provides a more comprehensive view of model performance across all classes.

\begin{equation}
    \text{Balanced Accuracy} = \frac{1}{5} \sum_{k=0}^{4} \frac{TP_k}{TP_k + FN_k}
    \label{eq:balanced_accuracy}
\end{equation}

\noindent Here, \( TP_k \), \( FP_k \), and \( FN_k \) correspond to the true positives, false positives, and false negatives for class \( k \), respectively. 

\subsection{AutoML Framework}
\label{sec:ml_framework}
In this study, we leveraged  AutoGluon \cite{autogluonl}, an open-source AutoML framework designed to automate the entire ML pipeline for tabular, time series and multimodal data.

Unlike traditional AutoML frameworks which primarily focus on model selection and hyperparameter optimization (e.g. CASH \cite{thornton_2013,zoller_2021} ), AutoGluon differentiates itself by prioritizing  advanced ensambling strategies.

\begin{figure}[H]
        \centering
        \includegraphics[width=0.5\linewidth]{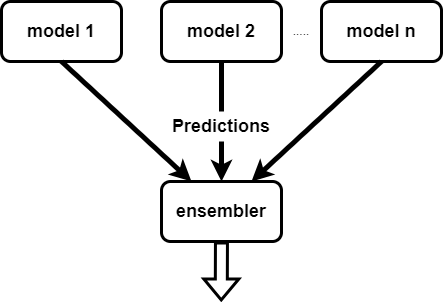}
        \caption{Ensembling Architecture}
        \label{fig:ensambling_architecture}
\end{figure}

In ensemble modeling, the predictions of multiple weak models are combined to build a single and more reliable outcome. Such approach is detailed in Figure \ref{fig:ensambling_architecture}.
Different models ($model_1, model_2, \ldots, model_n$) are trained independently on different chunks of the same dataset.
During inference, the predictions of the weak models are aggregated by the ensembler using techniques like majority voting for classification or averaging for regression.

\noindent Key innovations introduced by AutoGluon are:

\begin{itemize}
    \item \textbf{Multi-Layer Stack Ensembling}: This method constitutes a fully hierarchical ensembling approach that combines a set of diverse base models in a systematic way to increase both predictive accuracy and robustness. 
    More specifically, AutoGluon organizes models into hierarchical multiple layers of models where each layout can use the predictions from models in lower layers in the ensemble. 
    This hierarchical structure allows models in high layers to correct errors as well as capture meta-patterns that would be difficult to capture based upon the raw input data. 
    To reduce information loss and gain access to the original features, the hierarchical framework incorporates skip connections to feed raw input features into each layer models in the ensemble alongside the model predictions from lower layers.
    \item \textbf{Repeated k-Fold Bagging}: To reduce overfitting and improve robustness, AutoGluon decreases the variance of a model by training several instances of the model on distinct subsets of data. 
    The dataset is randomly split into k segments, where each model is trained using k-1 segments and validated on the remaining segment, thus producing out-of-fold (OOF) predictions. 
    These OOF predictions are then used by the framework to mitigate overfitting \cite{autogluonl}.
\end{itemize}

For additional information, the reader is directed to \cite{autogluonl}.

\subsection{SHAP}
SHAP (SHapley Additive exPlanations) \cite{lundberg2017unified} is a model-agnostic framework designed to provide consistent and grounded explanations for the output of any machine learning model. It is grounded in cooperative game theory and quantifies the importance of each input feature's contribution to a prediction using the Shapley values. SHAP provides local explanations (of individual predictions) and global explanations (of all predictions).

One popular global SHAP visualization is the feature importance plot, which summarizes the absolute SHAP values for all samples to determine the rank ordering of features by their total impact on the model's prediction. This visualization is a basic overview of the variables that influence the model's predictions overall. 

For a deeper-level analysis, the SHAP beeswarm plot shows the distribution of SHAP values across all instances in the dataset for each feature. Each dot is one observation and will generally be colored based on the actual value of that feature — red for high, blue for low. The positioning of the dots across the horizontal axis shows the SHAP value which describes the effect of that feature on the prediction of the model for that observation. The dots to the right of the mid-point (positive SHAP value) indicate the feature is positively affecting the prediction, while dots to the left indicate the feature is negatively affecting the prediction. This format allows one to see both the direction and the size of the influence a feature has, as well as how the values of the feature related to the variation in prediction.

\begin{figure}[h]
        \centering
        \includegraphics[width=0.7\linewidth]{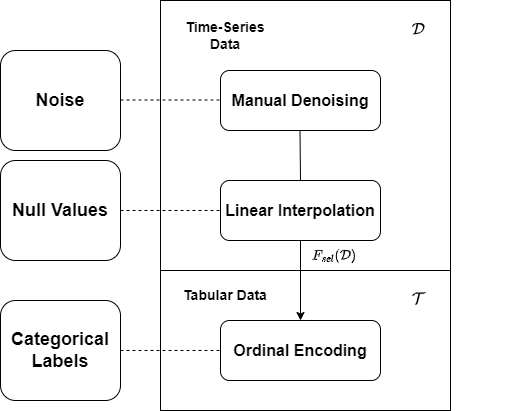}
        \caption{Preprocessing Operations}
        \label{fig:data_preprocessing.png}
\end{figure}

\begin{figure*}[h]
        \centering
        \includegraphics[width=0.7\linewidth]{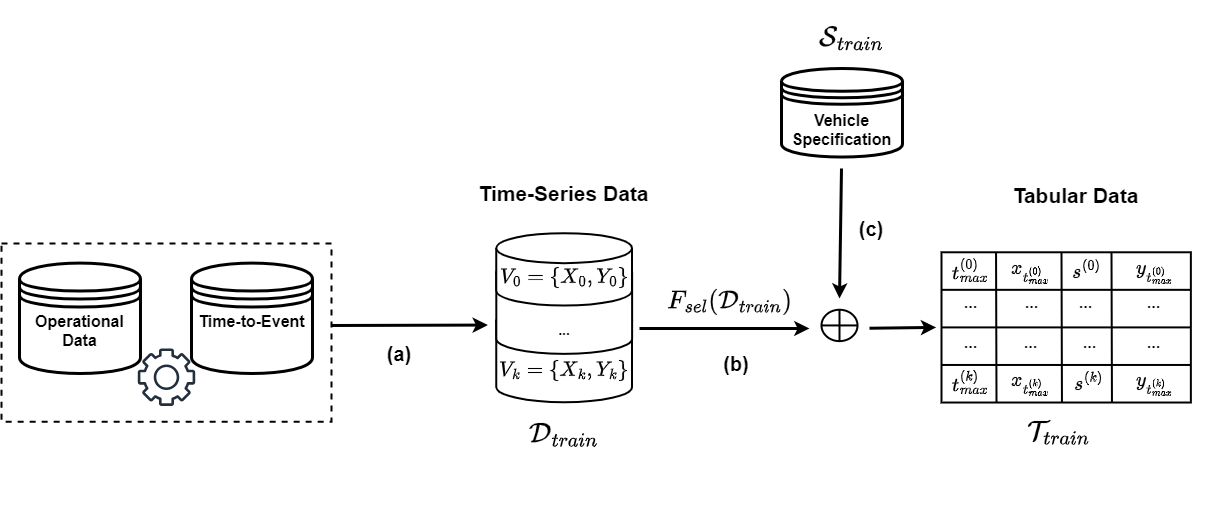}
        \caption{Data Transformation Pipeline For Training Partition}
        \label{fig:data_pipeline}
\end{figure*}
\section{\uppercase{Methodology}}

The  methodology adopted in this study is made up of three primary steps: (i) data wrangling, (ii) model training and (iii) performance evaluation.

The main idea underlying the methodology is to approximate the mapping function $f_{TS}$ defined in Equation~\ref{eq:mapping_function} using a tabular model $f_{TAB}$.
The approximation can be formalized as:
\begin{equation}
     f_{TS}(X_{i}) \approx f_{\text{TAB}}(f_{\text{sel}}(X_{i})), \quad \forall X_{i} \in \mathcal{D}
\end{equation}
where:
\begin{equation}
     X_{i} \in \mathbb{R}^{T_{i} \times M}, T_{i} \in \mathbb{N} 
\end{equation}
\begin{equation}
     f_{sel}: \bigcup_{T \in \mathbb{N}} \mathbb{R}^{T \times M} \rightarrow \mathbb{R}^{M}
\end{equation}
\begin{equation}
      f_{TAB}: \mathbb{R}^{M} \rightarrow \{0, 1, 2, 3, 4\}
\end{equation}

The first part of the approximation consists of the selection function $f_{\text{sel}}$ extracting the most recent observation from the time series $X_{i}$. 
In the second part, the tabular classifier $f_{\text{TAB}}$ assigns the selected observation to one of the discrete class labels.

% This method drew inspiration from the work of \cite{kargar-sharif-abad_shap-driven_nodate}, where the authors employed a similar approach on the same dataset to surrogate the explainability of survival models to regression models. However, while their selection method involves extracting a random observation from the time series, our methodology specifically focuses on the last observation. This choice ensures that the most recent state of the system is captured in the tabular dataset while avoiding further unbalancing toward class 0.

\subsection{Data Wrangling}

The data wrangling process began with correcting the operational readouts to satisfy the assumption of non-decreasing monotonicity, as illustrated in Figure~\ref{fig:data_preprocessing.png}. This involves removing noise types identified in Figure~\ref{fig:noise_types}.
Several denoising methods were explored, including average smoothing and outlier detection, but these methods proved insufficient presumably because the irregularity and multivariate nature of the data. Moreover, preserving the relationships between variables in the final observation was crucial for our methodology.
To address these challenges, we developed a custom script to systematically identify and correct anomalies: glitches were imputed to ensure a smooth transition back to normal values, while cumulative downward step noise was adjusted per single feature by adding the magnitude of each step to subsequent observations.

Regarding missing values, no feature was discarded, as each column in the operational data had less than 1\% of its values missing. 
The remaining missing values were imputed using linear interpolation, applied independently to each time series. 
Although the features do not exhibit perfectly linear patterns, this method is considered appropriate due to its simplicity and the minimal proportion of missing data. 

After denoising and imputation, we reduced the time-series as tabular data. Figure \ref{fig:data_pipeline} illustrates the data transformation process for the training partition. The operational readouts and time-to-event datasets were first transformed into the format \(\mathcal{D} = \{ V_i \}_{i=0}^{k}\), where \(V_i = (X_i, Y_i)\), by adding a new column called \textit{class\_label} to the operational readouts \(X\). 
Step (b) involved converting $\mathcal{D}$ into tabular form by extracting only the last observation from each series. In the final step (c), the output of \(F_{\text{sel}}\) was merged with the vehicle specifications dataset \(\mathcal{S}\).
The transformation applied to the validation and test partitions followed the same procedure, with the exception that the class labels were directly provided as part of the dataset, removing the need for the time-to-event auxiliary file.

At the end of the data transformation step we obtained three tabular datasets $\mathcal{T}_{train}$, $\mathcal{T}_{validation}$ and $\mathcal{T}_{test}$. 
Each data set contained a number of entries equal to the number of vehicles in the corresponding operational readout segments (23550, 5046, 5045) and 115 columns.

Subsequently, categorical variables were encoded using ordinal encoding, in which each unique category is mapped to an integer. This step is illustrated in the lower portion of Figure~\ref{fig:data_preprocessing.png}. Finally, the \texttt{vehicle\_id} column was dropped.

\begin{table*}[h]
    \centering
    \begin{tabular}{|l|c|c|c|c|}
        \hline
        Model & Balanced Accuracy (Test) & Challenge Metric (Test) & Balanced Accuracy (Val) & Challenge Metric (Val) \\
        \hline
        XGB & 0.2428 ± 0.01 & 39139.9 ± 1501.28 & 0.2411 ± 0.01 & 39331.75 ± 1094.43 \\
        GBM & 0.2375 ± 0.01 & 42193.71 ± 2348.89 & 0.25 ± 0.01 & 44655.71 ± 3168.26 \\
        CAT & 0.25 ± 0.01 & 38810.95 ± 821.01 & 0.2482 ± 0.01 & 41894.7 ± 2531.02 \\
        \hline
    \end{tabular}
    \caption{Aggregate Performance Comparison}
    \label{tab:aggregate_metrics}
\end{table*}

\subsection{Model Training}
We selected three families of ML models for our experiments: CatBoost, XGBoost, and LightGBM. 
The primary reason for choosing these gradient boosting (GB) algorithms is that they include built-in techniques to manage class imbalance, which is essential for our PdM task given the rarity of failure events, as noted in \cite{kharazian_2024}. 
Moreover, GB algorithms are inherently robust to outliers,  do not require feature normalization and capable of identifying non-linear connections between features and the target variables, making them particularly suitable for PdM tasks.

Since AutoGluon is deterministic, instead of performing hyperparameter optimization (HPO), each model was trained using 10 preconfigured configurations provided by AutoGluon. 
We leveraged AutoGluon's best quality preset, which offers a set of well-optimized hyperparameters designed to maximize predictive performance. 
This preset ensures a balance between model accuracy and computational efficiency, reducing the need for manual tuning.

Following model training, we evaluated the models on the validation dataset using the custom challenge metric. 
For each family, the final model selected was the one that minimized the challenge metric on the validation dataset.
Finally, the selected models were evaluated on the test dataset to assess their generalization performance.

\section{Results}

The results in Table~\ref{tab:aggregate_metrics} highlight variations in aggregate performance across models for both the test and validation datasets. The analysis, centered on the challenge cost metric introduced in Section \ref{sec:evaluation-metrics}, was conducted on 10 models from each model family.

\begin{table}[h!]
\centering
\caption{Best Model Results}
\begin{tabular}{|l|c|c|c|c|c|}
\hline
\textbf{Dataset} & \textbf{XGBoost} & \textbf{LightGBM} & \textbf{CatBoost} \\
\hline
Validation & \textbf{37406} & 40748 & 37595 \\
Test & 37733 & 42058  & \textbf{36724} \\
\hline
\end{tabular}
\label{tab:best_model_results}
\end{table}

Table~\ref{tab:best_model_results} presents the detailed challenge cost for the best models across training, validation, and test datasets.  While the XGB model achieved the lowest challenge cost on the validation dataset, the GBM model performed best on the training dataset. However, for a fair comparison with other baselines, XGB was selected as the best model.

\begin{table}[h]
    \centering
\begin{minipage}{0.45\textwidth}
    \caption{Confusion Matrix on Training Dataset}
    \vspace{4pt}
    \centering
    \begin{tabular}{|c|c|c|c|c|c|c|}
        \hline
        & \multicolumn{5}{|c|}{\textbf{Predicted}} & \\ \hline
        \multirow{6}{*}{\rotatebox{90}{\textbf{Actual}}} 
        & \textbf{0} & \textbf{1} & \textbf{2} & \textbf{3} & \textbf{4} & \\ \cline{2-7}
        & \textbf{0} & 17905 & 26   & 43   & 277  & 3037 \\ \cline{2-7}
        & \textbf{1} & 0     & 31   & 0    & 0    & 0    \\ \cline{2-7}
        & \textbf{2} & 0     & 0    & 70   & 0    & 0    \\ \cline{2-7}
        & \textbf{3} & 0     & 0    & 0    & 161  & 0    \\ \cline{2-7}
        & \textbf{4} & 108   & 0    & 0    & 18   & 1874 \\ \hline
    \end{tabular}
    \label{tab:confusion_matrix_validation}
\end{minipage}
    \begin{minipage}{0.45\textwidth}
        \caption{Confusion Matrix on Validation Dataset}
        \vspace{4pt}
        \centering
        \begin{tabular}{|c|c|c|c|c|c|c|}
            \hline
            & \multicolumn{5}{|c|}{\textbf{Predicted}} & \\ \hline
            \multirow{6}{*}{\rotatebox{90}{\textbf{Actual}}} && \textbf{0} & \textbf{1} & \textbf{2} & \textbf{3} & \textbf{4} \\ \cline{2-7}
            & \textbf{0} & 1839 & 11   & 7    & 28   & 3025 \\ \cline{2-7}
            & \textbf{1} & 5    & 0    & 0    & 0    & 11   \\ \cline{2-7}
            & \textbf{2} & 1    & 0    & 0    & 0    & 13   \\ \cline{2-7}
            & \textbf{3} & 6    & 0    & 0    & 0    & 24   \\ \cline{2-7}
            & \textbf{4} & 5    & 0    & 0    & 1    & 70   \\ \hline
        \end{tabular}
        \label{tab:confusion_matrix_validation}
    \end{minipage}
    \hspace{4pt}
    \begin{minipage}{0.45\textwidth}
        \caption{Confusion Matrix on Test Dataset}
        \vspace{4pt}
        \centering
        \begin{tabular}{|c|c|c|c|c|c|c|}
            \hline
            & \multicolumn{5}{|c|}{\textbf{Predicted}} & \\ \hline
            \multirow{6}{*}{\rotatebox{90}{\textbf{Actual}}} && \textbf{0} & \textbf{1} & \textbf{2} & \textbf{3} & \textbf{4} \\ \cline{2-7}
            & \textbf{0} & 1756 & 148  & 4    & 67   & 2928 \\ \cline{2-7}
            & \textbf{1} & 2    & 1    & 0    & 1    & 22   \\ \cline{2-7}
            & \textbf{2} & 2    & 0    & 0    & 1    & 12   \\ \cline{2-7}
            & \textbf{3} & 2    & 0    & 0    & 0    & 39   \\ \cline{2-7}
            & \textbf{4} & 8    & 1    & 0    & 0    & 51   \\ \hline
        \end{tabular}
        \label{tab:confusion_matrix_test}
    \end{minipage}
\end{table}

\noindent Tables \ref{tab:confusion_matrix_test} and \ref{tab:confusion_matrix_validation} show the confusion matrices of the best-performing model, XGB, on the test and validation datasets. It can be observed that the model struggles to correctly classify observations belonging to the intermediate classes (1, 2, and 3). Further details about this aspect will be given in the next section.

\begin{table}[h]
    \caption{State-Of-The-Art Results}
    \centering
    \begin{tabular}{|c|c|c|}
        \hline
        \textbf{Study Reference} & \textbf{Model} & \textbf{Test Cost} \\
        \hline
        \textbf{Dimidov et al. (Our approach)} & \textbf{XGBoost} & \textbf{37733} \\
        Zhong and Wang \cite{zhong} & Bi-LSTM & 39123 \\
        Parton et al. \cite{parton} & GNN & 47612 \\
        Carpentier et al. \cite{carpentier} & XGBoost & 49671 \\
        \hline
    \end{tabular}
    \label{tab:sota}
\end{table}

Overall, the performance of our methodology is competitive with respect to the current SOTA results. As shown in Table~\ref{tab:sota}, our approach improves on previous work by reducing the test cost from 39K to 37K. Furthermore, our method is computationally lightweight

Notably, in the case of \cite{carpentier}, the XGBoost model was evaluated using a different feature set, extracted through a windowing approach combined with the tsfresh library. In contrast, our method selects observations more effectively, ensuring that the model captures the most relevant patterns, which likely explains the improved performance.

\section{Model Explainability}
While achieving good predictive performance is crucial in PdM applications for economical reasons, it is equally important that the decision-making process of the models remains interpretable. In industrial contexts, engineers and fleet managers need to understand the underlying reasoning for each maintenance suggestion to promote accountability and build trust in ML-based tools. In this section, we provide insights into the inner workings of our best-performing model by analyzing feature importance and explaining individual predictions using SHAP values.

\subsection{Global Explanation}

\begin{figure*}[h]
        \centering
        \includegraphics[width=0.7\linewidth]{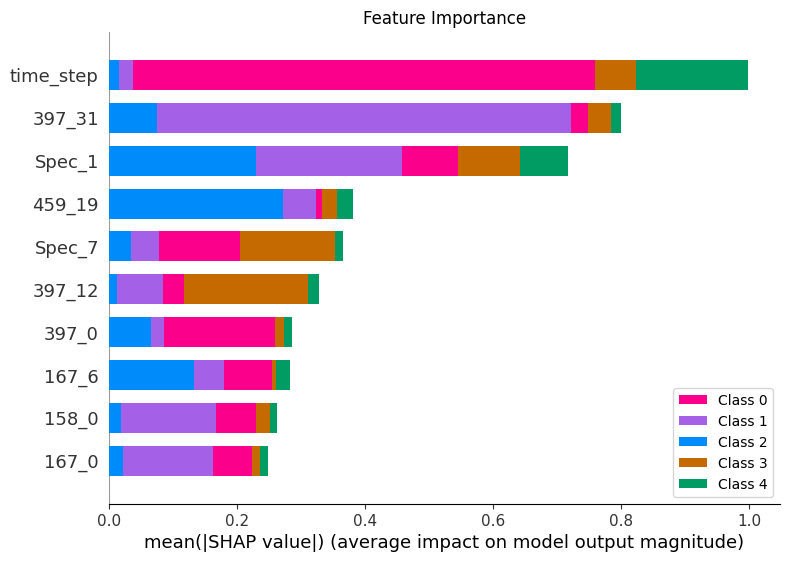}
        \caption{Global Feature Importance}
        \label{fig:global_feature_importance}
\end{figure*}

Figure \ref{fig:global_feature_importance} illustrates the global feature importance of the final model, aggregated across all classes using SHAP values. Among all features, \texttt{time\_step} stands out as the most influential predictor, with substantial impact on class 0 and class 4. In addition, sensor-derived features such as \texttt{397\_31}, \texttt{459\_19}, and categorical vehicle specifications like \texttt{Spec\_1} and 
\texttt{Spec\_7} also play an important role.

\subsection{Class-Specific Insights}

\begin{figure*}[ht]
    \centering

    % First row: Class 1, 2, 3

    \begin{subfigure}[b]{0.3\textwidth}
        \includegraphics[width=\textwidth]{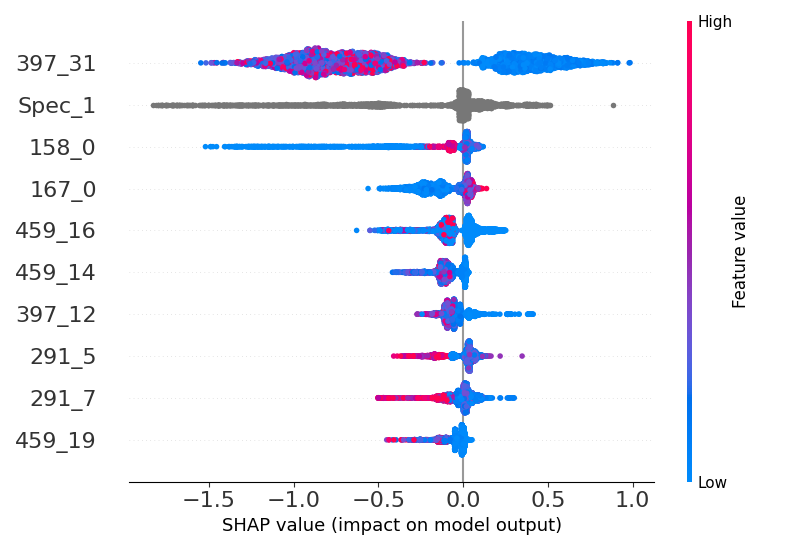}
        \caption{Class 1}
        \label{fig:shap_class1}
    \end{subfigure}
    \hfill
    \begin{subfigure}[b]{0.3\textwidth}
        \includegraphics[width=\textwidth]{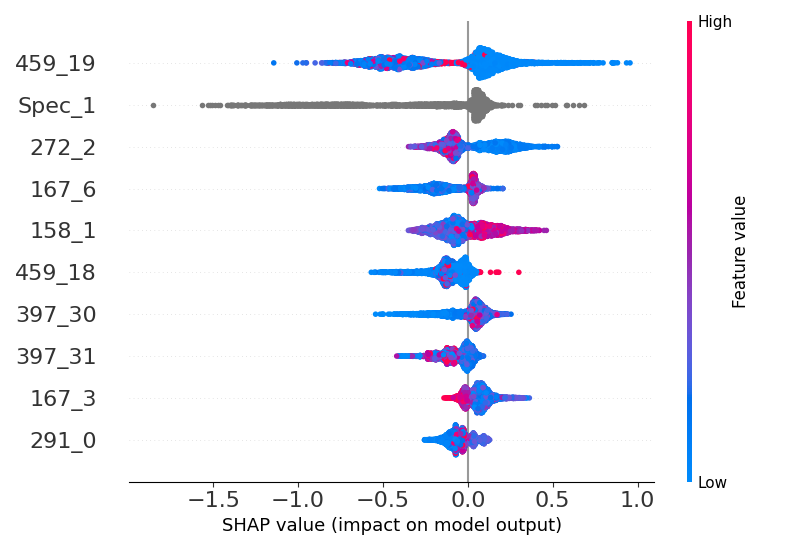}
        \caption{Class 2}
        \label{fig:shap_class2}
    \end{subfigure}
    \hfill
    \begin{subfigure}[b]{0.3\textwidth}
        \includegraphics[width=\textwidth]{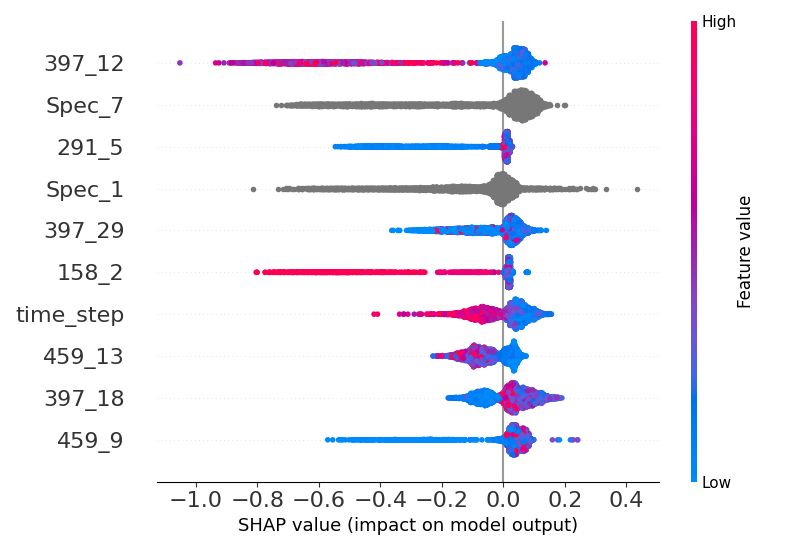}
        \caption{Class 3}
        \label{fig:shap_class3}
    \end{subfigure}
    \hfill

    \vspace{0.5cm}

    % Second row: Class 0 and Class 4 centered
    \hspace*{0.15\textwidth}
    \begin{subfigure}[b]{0.3\textwidth}
        \includegraphics[width=\textwidth]{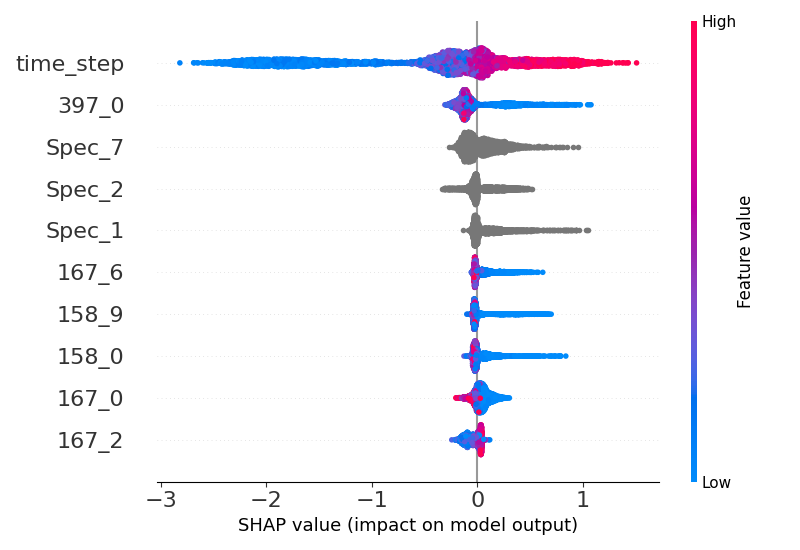}
        \caption{Class 0}
        \label{fig:shap_class0}
    \end{subfigure}
    \hfill
    \begin{subfigure}[b]{0.3\textwidth}
        \includegraphics[width=\textwidth]{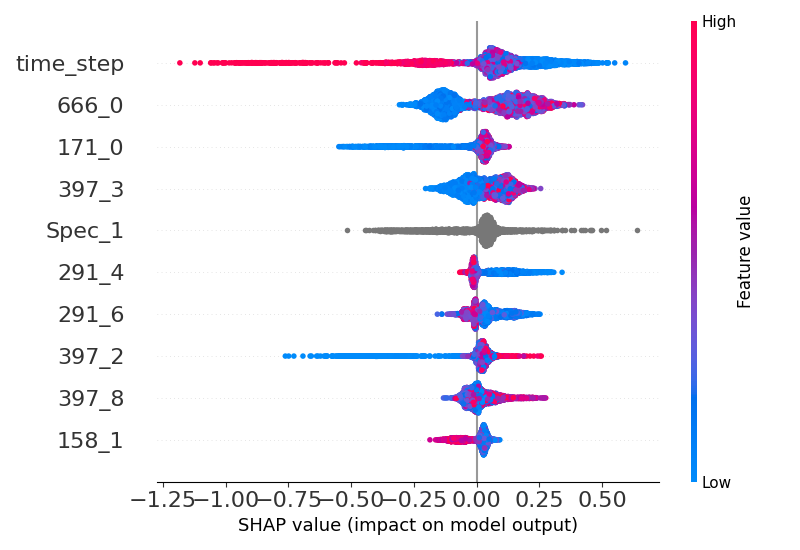}
        \caption{Class 4}
        \label{fig:shap_class4}
    \end{subfigure}

    \caption{Class-Specific Feature Importance.}
    \label{fig:shap_per_class}
\end{figure*}

To enhance our comprehension of the model's behavior for each class, we examined SHAP values with beeswarm plots across different degradation categories (Figures \ref{fig:shap_per_class})
\begin{itemize}
    \item Class 0 (Figure d): High shap values for feature \texttt{time\_step} push predictions toward the healthy class. Other features like \texttt{397\_0} and \texttt{158\_0} also support class 0 when they have low values, indicating minimal component stress and early usage phases.
    \item Class 4 (Figure e): Medium-to-low \texttt{time\_step} values strongly increase the probability of imminent failure, as do high values in \texttt{666\_0}, \texttt{397\_3}, \texttt{397\_2} and \texttt{397\_8}. These indicate accumulated stress and usage. However, there are features like \texttt{291\_4} and \texttt{291\_6} which contribute positively to the class through low values. 
    \item Classes 1-3 (Figures a–c): The intermediate levels are identified by a combination of different features, with no single dominant pattern. Interestingly, the \texttt{time\_step} feature plays a less prominent role compared to classes 0 and 4. Instead, the model relies more on histogram-based operational signals like \texttt{397\_31}, \texttt{459\_19}, and \texttt{272\_2}. The mixed features and the absence of patterns reflect the ambiguity in these classes, making them harder to distinguish reliably.
\end{itemize}

\section{Discussion}
\label{sec:discussion}

% limitations of the methodology
% - prediction horizon da 48 time\_step a 6 time\_step -> less time to order the component in advance
% - requires more cost for doing remote monitoring
% strengths of the methodology
% + less storage costs
% riflessione sulla funzione di costo: cost function che tende a minimizzare i guasti inaspettati.

The findings of this investigation suggest that proper modeling of component wear-and-tear, failure processes, as well as wise observation selection could assist with the creation of a viable PdM framework for the automotive world. 

\subsection{Methodology Applicability}
\label{model_applicability}

We believe this approach is an appropriate methodology for automotive parts that exhibit cumulative and irreversible deterioration. 
For example, engine and powertrain components (turbochargers, fuel injection systems, and cylinders) suffer gradual wear due to erosion, contamination, and friction. 
Similarly, wear on the transmission and drivetrain components, like clutches, differentials, and driveshafts, persists if they are not adequately serviced. 
Finally, further notable examples are braking and suspension components (brake pads, shock absorbers, and wheel bearings), which degrade over time due to endless miles traveled, sudden stops, and bad road conditions.

In order to quantify the usage of these automotive parts, a variety of sensor-derived features can be integrated into the predictive model. 
Some signals never decrease by nature. 
For example, odometer readings and engine hours, which may indicate general wear and usage progression for many components. 
Other signals fluctuate within a limited range and may be converted into increasing features through binning and counting. 
For instance, engine RPM in trucks can be categorized into ranges such as \{(RPM< 1400), (RPM $\geq$ 1400)\}. 
Counting how often the RPM falls within each bin when the vehicle is operative will provide cumulative metrics that reflect usage patterns over time.

\subsection{Methodology Limitations}

Representing the wear state in terms of a monotonic, non-decreasing feature set has some drawbacks. 
In particular, this modeling approach is unlikely to generalize well for components with cyclic or environment-dependent degradation patterns. 
As an example, batteries degrade non-linearly, with temporary performance recovery after recharging, while environmental factors (e.g., temperature) introduce additional variability. 

In addition, transforming time-series into tables while only considering the most recent observation can remove important temporal information. 
Some types of failures are predictable not by the wear-and-tear state but rather based on the evolution of signals over time. 
By ignoring the time context, the model will not learn how to spot transient anomalies or sudden degradation by the model.

Another empirical limitation is that the tabular models tends to correctly classify only the observations belonging to categories 0 and 4. 
This suggests that such models have a short prediction horizon, as they face challenges in forecasting distant failures (classes 1, 2 and 3), but are highly effective at detecting failures that are about to occur (class 4). 
Moreover, the presence of many false positives degrades the PdM policy into a kind of preventive maintenance (PM). 
Notably, the challenge did not specify the minimum cost threshold at which a fleet manager would prefer transitioning from a PM to a PdM policy. 
Besides, the challenge metric appears to prioritize failure detection over the ability to spot long-term degradation, as it imposes significantly higher penalties on false negatives associated with class 4.

\subsection{Methodology Justification}

Deep learning models, like RNNs and Transformers, can learn temporal dependencies and model dynamics directly from raw sequences, without making any assumptions on the feature trends. 
The top performing method in [7] utilized Bi-LSTM models to capture both forward and backward temporal patterns and perhaps exploited weak degradation signals at early time points that our models are not capable to exploit. 
The shortcoming of the DL approaches is that they are computationally expensive, needs large volume of data for training and often lack interpretability, a concern increasingly in demand in industry.
So, while our tabular approach offers simplicity, interpretability, and computational efficiency, making it suited for practical work environments, the compromise may result in a lower generalization and predictive power.

\subsection{Real World Deployment and Scalability}

The proposed methodology is well-suited for real-world deployment in the automotive fleet industry for several reasons.

Firstly, the method trains ML models on a tabular dataset of size $\mathcal{O}(N\cdot M)$. Since $M$ is fixed and relatively small, the space requirements scale linearly with the size of the fleet. Regarding time complexity, the methodology requires to train a constant number of GB models, where training a single one cost approximately $\mathcal{O}(N\cdot M\cdot \log N)$. Concerning inference time, predictions for new vehicles involve traversing a small number of decision trees, which is roughly $\mathcal{O}(Td)$, where $T$ is the number of trees and $d$ is their average depth. 
We believe that these properties make the approach computationally efficient and scalable to fleets comprising tens of thousands of vehicles.

Secondly, the use of tabular models, such as XGBoost, and AutoGluon’s AutoML framework, enables straightforward integration with existing cloud ML platforms, facilitating rapid deployment and automated updates. The system can be deployed as a lightweight service, which consumes the most recent operational readout from a vehicle and returns a degradation score. Subsequently, this service can be integrated with dashboards and visualization platforms.

However, the long-term practicability of the solution presents some downsides. It does not take into account the dynamic nature of the fleets and their operating conditions. For this reason, the model must be periodically retrained to account for changes in fleet composition and usage patterns. 
Additionally, although the methodology supports fast inference on individual observations, real-time constraints in safety-critical environments may necessitate stricter latency limits and improved streaming functionalities. 
\section{Conclusions and Future Directions}
\label{sec:conclusion}

% state that the approach is suitable for MVP

This paper presents a methodology for converting a time-series PdM task into a tabular one, under the constrain that the time-series are monotonically non-decreasing.
The methodology is implemented within a pipeline that integrates an AutoML tool and is benchmarked against SOTA approaches using the Scania dataset. Experimental results indicate that the proposed approach is competitive, offering a straightforward and cost-effective alternative for developing PdM systems, especially in resource constrained settings.

The current evaluation is limited to the Scania dataset. To further validate the methodology, additional experiments on diverse fleets and components are necessary. Accordingly, future work will focus on applying the approach to new, non-anonymized use cases, following the guidelines discussed in Section \ref{model_applicability}. Another promising research direction involves extending the methodology to hybrid models in order to overcome limitations related to short prediction horizons and many false positive observed in the results. Finally, further investigation could focus on integrating the methodology with high-throughput data pipelines and deploying it on edge infrastructure to assess its suitability to scenarios requiring strict latency and high throughput.

Ultimately, this methodology opens the door to more accessible and scalable PdM solutions, bridging the gap between time-series data and efficient model deployment.

\section{Acknowledgements}
\label{sec:acknowledgments}

This research was funded in whole, or in part, by the Luxembourg National Research Fund (FNR), grant reference BRIDGES/2022/IS/17270233. For the purpose of open access, and in fulfillment of the obligations arising from the grant agreement, the author has applied a Creative Commons Attribution 4.0 International (CC BY 4.0) license to any Author Accepted Manuscript version arising from this submission.

\bibliographystyle{IEEEtran}
\bibliography{mybibfile}

% that's all folks
\end{document}